\pdfoutput=1
\documentclass[9pt,twocolumn,twoside]{IEEEtran}
\usepackage[colorlinks=true]{hyperref}

\usepackage{cite,amssymb,multirow,fixltx2e,graphicx}
\usepackage[cmex10]{amsmath}
\ifCLASSOPTIONtwocolumn
\newcommand{\twocm}[1]{#1}
\newcommand{\onecm}[1]{}
\else
\newcommand{\twocm}[1]{}
\newcommand{\onecm}[1]{#1}
\fi

\title{Image Restoration with\onecm{\\} Signal-dependent Camera Noise}%
\author{Ayan Chakrabarti\onecm{$^*$} and Todd Zickler\onecm{\vspace{-4em}}%
\thanks{The authors are with the School of Engineering and Applied Sciences, Harvard University, Cambridge, MA, 02138 USA (e-mail: \{ayanc,zickler\}@eecs.harvard.edu).}}%

\begin{document}
\maketitle

\begin{abstract}
This article describes a fast iterative algorithm for image denoising and deconvolution with signal-dependent observation noise. We use an optimization strategy based on variable splitting that adapts traditional Gaussian noise-based restoration algorithms to account for the observed image being corrupted by mixed Poisson-Gaussian noise and quantization errors.
\end{abstract}


\section{Introduction}
\label{sec:intro}

\IEEEPARstart{I}{mage} restoration refers to the recovery of a clean sharp image from a noisy, and potentially blurred, observation. Most state-of-the-art restoration techniques~\cite{portilla,bm3d,levin07,ADM,dilip} assume that observed image is corrupted by signal-independent additive white Gaussian noise (AWGN), since this makes both analysis and estimation significantly more convenient. However, observation noise in an image captured by a real digital camera is typically signal-dependent, and therefore image restoration algorithms based on AWGN models make sub-optimal use of the information available in the observed image. For example, one source of noise in recorded intensities is the uncertainty in the photon arrival process, which is Poisson distributed and has a variance that scales linearly with the signal magnitude. Therefore, using an AWGN model fails to account for the noise variance at darker pixels being lower than that at brighter ones, and can lead to over-smoothing in darker regions of the restored image.

Due to these reasons, the development of restoration algorithms based on accurate noise models has been an area of active research~\cite{kolaczyk, hirakawa, florian, foi0, foi}. However, this task is made challenging by the fact that state-of-the-art restoration methods use sophisticated image priors that are defined in terms of coefficients of some spatial transform, and combining these priors with a non-Gaussian noise model significantly adds to complexity during estimation. Denoising methods for signal-dependent noise are either based on a (sometimes approximate) statistical characterization of this noise in transform coefficients~\cite{hirakawa,kolaczyk,florian}, or use a \emph{variance-stabilization} transform that renders the noise Gaussian followed by traditional AWGN denoising techniques~\cite{foi0,foi}. Recently, Harmany~et al.~\cite{spiraltv} presented an iterative deconvolution algorithm that approximates a Poisson-noise likelihood cost at each iteration with a quadratic function based on the curvature of the likelihood at the current estimate. This technique was combined with various sparsity-based priors in \cite{spiraltv} to enable restoration in the presence of Poisson noise.

We describe an iterative image restoration framework that accounts for the statistics of real camera noise. It is also an iterative technique like the one in \cite{spiraltv}, but uses an optimization strategy known as \emph{variable-splitting} leading to significant benefits in computational cost. The use of this strategy for image restoration dates back to the work of Geman et al.~\cite{geman}, and has been used to develop fast deconvolution algorithms that use the so-called \emph{total-variation} (TV) image prior model~\cite{ADM}, with extensions that consider non-quadratic penalties on the noise--- including a Poisson likelihood cost~\cite{uclatr}. In this paper, we deploy this technique to enable efficient estimation with a likelihood cost that models observation noise as a combination of Gaussian noise, signal-dependent Poisson shot-noise, and digitization errors. Moreover, we develop a general framework that is not tied to any specific choice of image prior, and this allows us to adapt any state-of-the-art AWGN restoration technique (such as BM3D~\cite{bm3d} for denoising) for use with the proposed signal-dependent noise model. We demonstrate the efficacy of this approach with comparisons to both AWGN and signal-dependent state-of-the-art restoration methods.

\section{Observation Model}
\label{sec:nmodel}

Let $x(n)$ be the latent noise-free sharp image of a scene corresponding to an observed image $y(n)$, where $n \in \mathbb{R}^2$ indexes pixel location. We assume that a spatially-uniform blur acts on the scene, with a known kernel $k(n)$. We let $x_k(n) = (x*k)(n)$ denote the blurred image that would have been observed in the absence of noise. The observation $y(n)$ can then be modeled as
\begin{equation}
  \label{eq:sensor}
  y(n) = Q\left[\tilde{y}(n)\right],~~\tilde{y}(n) = y_k(n) + z(n),
\end{equation}
where $Q(\cdot)$ is a quantization function used to digitize the analog sensor measurement $\tilde{y}(n)$, which we in turn model as the sum of a scaled Poisson random variable $y_k(n)$ with mean $x_k(n)$, and zero-mean Gaussian random noise $z(n)$ with variance $\sigma^2$. We examine the statistical properties of each component of the model in \eqref{eq:sensor}, and define a likelihood function based on these properties with the aim of enabling accurate yet efficient inference of $x(n)$ from $y(n)$.

\subsection{Shot Noise}
\label{sec:shotnoise}

At each location $n$, the random variable $y_k$  captures the uncertainty in the photon arrival process, and is modeled with a scale parameter $\alpha > 0$ as $\alpha y_k \sim \mathcal{P}(\alpha x_k)$, i.e.,
\begin{equation}
  \label{eq:pzn}
  P\left(y_k | x_k\right) = \frac{\left(\alpha x_k\right)^{\alpha y_k}e^{-\alpha x_k}}{\left(\alpha y_k\right)!}.
\end{equation}
The difference between the observed $y_k$ and its mean $x_k$ is referred to as shot noise, and has a signal-dependent variance equal to $x_k/\alpha$. The parameter $\alpha$ depends on the ISO setting of the camera, and a high value of $\alpha$ corresponds to a low ISO setting and a higher signal-to-noise ratio (SNR).  We define $L_P(x_k;y_k,\alpha)$ as the corresponding negative log-likelihood (up to a constant) of the observed pixel $y_k$ being due to $x_k$:
\begin{equation}
  \label{eq:Lpz}
  L_P(x_k;y_k,\alpha) = \alpha x_k - \alpha y_k\log x_k.
\end{equation}
Note that $L_P$ is a convex function of $x_k$ (since $\partial^2L_P/\partial x_k^2=\alpha{}y_k/x_k^2\geq 0, \forall x_k$), with a minimum at $x_k = y_k$.

\subsection{Gaussian Noise}
\label{sec:gnoise}

In addition to shot noise, the measurement $\tilde{y}$ may be corrupted by other signal-independent noise sources, such as thermal and amplifier noise, which we model with the AWGN variable $z \sim \mathcal{N}(0,\sigma^2)$~\cite{foi,florian}. Combining this with the Poisson model in \eqref{eq:pzn}, we have:
\begin{equation}
  \label{eq:pmdl}
  p\left(\tilde{y}|x_k\right) \propto \sum_{r=0}^\infty \frac{(\alpha x_k)^re^{-\alpha x_k}}{r!} \exp\left(- \frac{(\tilde{y} - r/\alpha)^2}{2\sigma^2} \right).
\end{equation}
Unfortunately, the above expression can not be computed in closed form, and therefore we employ an approximation to define the log-likelihood function for this case. We note that $\tilde{y}$ is a mixed Poisson-Gaussian random variable with mean $x_k$ and variance $(x_k/\alpha+\sigma^2)$, and approximate it with a shifted Poisson likelihood as
\begin{equation}
  \label{eq:LGpsz}
  L_{PG}(x_k;\tilde{y},\alpha,\sigma) = L_P(x_k + \alpha\sigma^2;\tilde{y} + \alpha\sigma^2;\alpha).
\end{equation}

\subsection{Quantization}
\label{sec:quant}

Finally, we account for the observed intensity $\tilde{y}$ being quantized by a function $Q(\cdot)$ that maps every interval $[y_-,y_+]$ to a single value $y$. We consider the general case where quantization is possibly preceded by a non-linear map for gamma-correction:
\begin{equation}
  \label{eq:quant1}
  y = \left(\left\lfloor \tilde{y}^{1/g} \right\rfloor_q \right)^g,
\end{equation}
where $g$ corresponds to the gamma-correction exponent ($1$ for linear data, and typical $2.2$ for sRGB), and $\lfloor\cdot\rfloor_q$ denotes rounding off intervals of width $q$. Note that $y$ here denotes a linearized version of the camera image, i.e., one where the inverse of the gamma-correction function has been applied to the quantized observations. The interval $[y_-,y_+]$ may therefore be asymmetric around $\tilde{y}$ and have variable widths, and the mean and variance of $\tilde{y}$ given $y$ are given by
\begin{eqnarray}
  \label{eq:quant2}
  m_q(y) \hspace{-0.5em}&=\hspace{-0.5em}& \mathbb{E}[\tilde{y}|y] = \frac{\left(y^{1/g}+q/2\right)^{g+1}-\left(y^{1/g}-q/2\right)^{g+1} }{q(g+1)},\notag\\
  \sigma_q^2(y)\hspace{-0.5em}&=\hspace{-0.5em}&\mathbb{E}[(\tilde{y}-m_q(y))^2|y]
\twocm{\notag\\&=\hspace{-0.5em}&}\onecm{=}
\frac{\left(y^{1/g}+q/2\right)^{2g+1}-\left(y^{1/g}-q/2\right)^{2g+1} }{q(2g+1)}-m^2_q(y).~~~~
\end{eqnarray}
Note that this variance is only signal-dependent when $g\neq 1$, since for  $g=1$,  $\sigma_q^2(y)=q^2/12$ is independent of $y$. We incorporate these to obtain the overall likelihood function as
\begin{eqnarray}
  \label{eq:LPGQ}
  L(x_k;y,\alpha,\sigma,Q)=L_{PG}(x_k;m_q(y),\alpha,\sqrt{\sigma^2+\sigma_q^2(y)}).
\end{eqnarray}

\section{MAP-based Restoration}
\label{sec:algo}

With a suitably defined prior model $p(x)$ for natural images, we recover the maximum a-posteriori (MAP) estimate $\hat{x}(n)$ of $x(n)$ from $y(n)$ as:
\begin{equation}
  \label{eq:map}
  \hat{x} = \arg \min_x \Phi(x) + \sum_n L((x*k)(n);y(n)),
\end{equation}
where $\Phi(x)=-\log p(x)$, and $L(\cdot)$ is the likelihood function defined in \eqref{eq:LPGQ} (with the arguments $\alpha,\sigma,Q$ omitted for brevity). 

The main obstacle to computing a solution for \eqref{eq:map}, even in the absence of blur (i.e., $k = \delta$), is the fact that natural image priors are best defined in terms of coefficients in some transform domain. Unlike the AWGN case where noise in the coefficients of the observed image is also independent and Gaussian, estimation under the noise model in \eqref{eq:LPGQ} is challenging because of the complexity in characterizing the statistics of noise in the transform domain. 

\subsection{Variable Splitting}
\label{sec:vsplit}

We use an optimization approach similar to the one in \cite{uclatr}, that allows us to deal with minimizing the prior and likelihood terms in their respective \emph{natural} domains, i.e., in terms of transform coefficients and individual pixels respectively. Specifically, we recast the unconstrained minimization problem in \eqref{eq:map} as an equality-constrained optimization with the addition of new variables $\tau(n)$:
\begin{eqnarray}
  \label{eq:map2}
  \hat{x} = \arg \min_x \Phi(x) + \sum_n L(\tau(n);y(n)),\twocm{\notag\\}
\mbox{~~~subject to:~} \tau(n) = (x*k)(n).
\end{eqnarray}
Clearly, the problems in \eqref{eq:map} and \eqref{eq:map2} are equivalent. However, this modified formulation can be solved using an efficient iterative approach that allows the noise in each pixel to be treated independently.

Note that instead of choosing a specific image prior, we assume the existence of a baseline AWGN-based image restoration algorithm that (perhaps implicitly) defines $\Phi(x)$, and also provides an estimator function $G(\cdot)$:
\begin{equation}
  \label{eq:baseline}
   G(y,k,\sigma) = \arg \min_x \Phi(x) + \sum_n \frac{\left[y(n)-(x*k)(n)\right]^2}{2\sigma^2}.
\end{equation}
While we treat $G(\cdot)$ as a ``black box'' in general, the appendix describes a ``parallel'' optimization algorithm for a special case when the baseline algorithm is itself based on variable-splitting.

\subsection{Minimization with Augmented Lagrangian Multipliers}
\label{sec:alm}

We use the augmented Lagrangian multiplier method~\cite{uclatr} to solve this constrained optimization, and define a new augmented cost function that incorporates the equality constraint:
\begin{eqnarray}
  \label{eq:alm}
  \hspace{-1em}C(x,\tau,\lambda)\hspace{-0.6em}&=\hspace{-0.6em}&\Phi(x) + \left[ \frac{\beta}{2}\sum_n (\tau(n)-x_k(n))^2
\twocm{\right.\notag\\&&\hspace{-4em}\left.} 
- \sum_n \lambda(n)\left(\tau(n)-x_k(n)\right) \right] + \sum_n L(\tau(n);y(n)),
\end{eqnarray}
where $x_k(n) = (x*k)(n)$, $\lambda(n)$ are the Lagrange multipliers, and $\beta$ is a fixed scalar parameter. Note that the Lagrangian terms in this expression are augmented with an additional quadratic cost on the equality constraint. This additional cost is shown to speed up convergence, and since it has a derivative equal to zero when the equality constraint is satisfied, it does not affect the Karush-Kuhn-Tucker (KKT) condition at the solution.

The solution to \eqref{eq:map2} can be reached iteratively by making the following updates to $x(n),\tau(n)$ and $\lambda(n)$ at each iteration:
\begin{eqnarray}
  \label{eq:update1}
  x^{t+1} &\leftarrow& \arg \min_x C(x,\tau^t,\lambda^t),\\
  \label{eq:update2}
  \tau^{t+1} &\leftarrow& \arg \min_\tau C(x^{t+1},\tau,\lambda^t),\\
  \label{eq:update3}
  \lambda^{t+1}(n) &\leftarrow& \lambda^t(n) - \gamma(n) \beta (\tau^{t+1}(n)-x_k^{t+1}(n)),
\end{eqnarray}
where $x^t,\tau^t,\gamma^t$ refers to the values of these variables at iteration $t$, and the step size $\gamma(n)$ lies between $0$ and $\gamma_{\mbox{\tiny max}}=(\sqrt{5}+1)/2$. 

The update to $x$ in \eqref{eq:update1} involves minimizing the sum of the prior term $\Phi(x)$ with a uniformly-weighted quadratic cost, and  this can be achieved using the baseline estimator $G(\cdot)$ as
\begin{equation}
  \label{eq:xupd}
  x^{t+1} = G\left(\tau^{t}(n)-\beta^{-1} \lambda^{t}(n),k,\beta^{-1}\right).
\end{equation}
The update to $\tau$ in \eqref{eq:update2} involves a minimization that is independent of the prior term, and can be done on a per-pixel basis. For each $n$, we solve for $\partial C/\partial \tau(n) = 0$ to obtain
\begin{eqnarray}
  \label{eq:tausol}
  &
  \tau^{t+1}(n) = (b+\sqrt{b^2+4c})/2,\notag\\
  &\hspace{-2.6em}
  b=\left(x^{t+1}_k(n) +\beta^{-1}\lambda^t(n)\right)- \alpha\beta^{-1} - \alpha\left(\sigma^2+\sigma_q^2(y(n))\right),\notag\\
  &\hspace{-1.5em}
c=\alpha\left(\sigma^2+\sigma_q^2(y(n))\right)\left(x_k^{t+1}(n)+\beta^{-1}\lambda^t(n)\right) + \alpha\beta^{-1} m_q(y(n)).\hspace{-1.5em}\twocm{\notag\\}
\end{eqnarray}

\subsection{Algorithm Details}
\label{sec:details}

We begin the iterations with $\tau(n) = m_q(y(n))$ and $\lambda(n) = 0$, and stop when the relative change in $x$ falls below a threshold:
\begin{equation}
  \label{eq:thresh}
  \frac{\|x^{t+1}(n)-x^{t}(n)\|^2}{\|x^{t}(n)\|^2} \leq \epsilon.
\end{equation}
We find that it is optimal to vary the step-size $\gamma(n)$ at each pixel (but this is kept fixed across iterations), with a higher step-size for pixels with higher observed intensities $y(n)$ that have a higher expected noise variance. Specifically, we vary the step-size linearly with respect to $y(n)$, between $\gamma_{\mbox{\tiny max}}/2$ and $\gamma_{\mbox{\tiny max}}$.

Finally, the choice of $\beta$ involves a trade off between accuracy and speed of convergence. We find that it is best to choose a value inversely proportional to the average expected noise variance in the image, i.e. $\beta=\beta_0 / \sigma_{\mbox{avg}}^2$, where
\begin{equation}
  \label{eq:betachoice}
  \sigma_{\mbox{avg}}^2= \alpha^{-1}\mbox{avg}\{m_q(y(n))\} + \sigma^2 + \mbox{avg}\{\sigma_q^2(y(n))\},
\end{equation}
and $\beta_0$ depends on the choice of the baseline algorithm $G(\cdot)$.

\section{Experimental Results}
\label{sec:exp}

We use synthetically blurred and noisy images to compare the proposed approach to traditional AWGN-based methods for deconvolution, as well as Poisson-Gaussian noise-based methods for denoising~\cite{foi,florian}. Table~\ref{tab:dcnv} shows deconvolution performance on three standard images blurred with circular pill-box kernels (that correspond to typical defocus blur) of different radii $r$, in the presence of Poisson-Gaussian noise with different values of $\alpha$ and $\sigma$, as well as  with quantization errors (marked as $+Q$ in the table) corresponding to 8-bit quantization post gamma-correction ($q=1/256,g=2.2$). 

\begin{table*}[!t]
\twocm{\renewcommand{\arraystretch}{1.3}}
\onecm{\renewcommand{\arraystretch}{0.9}}
  \centering
  \caption{Deconvolution performance (PSNR in dB) for Poisson-Gaussian and Quantization\hspace{25em} Noise (indicated with +Q), and blur with circular pill-box kernels of radius $r$}
  \label{tab:dcnv}

{\onecm{\tiny}
  \begin{tabular}{|c|c|c||c|c|c|c|c|c|c|c|c|c|c|c|c|}
    \hline
    \multirow{2}{*}{$\alpha$} & 
    \multirow{2}{*}{$\sigma$} & 
    \multirow{2}{*}{Method} & 
    \multicolumn{4}{|c|}{Cameraman} &
    \multicolumn{4}{|c|}{Lena} &
    \multicolumn{4}{|c|}{Boats}\\\cline{4-15}
    &&&
    $r=5$&$r=7$&$r=9$&$r=11$&
    $r=5$&$r=7$&$r=9$&$r=11$&
    $r=5$&$r=7$&$r=9$&$r=11$\\\hline\hline

\multirow{3}{*}{$1024$}&
\multirow{3}{*}{$10^{-4}$}&
Input&
20.55&
19.46&
18.64&
17.97&
25.02&
23.62&
22.58&
21.76&
23.12&
22.05&
21.35&
20.81\\\cline{3-15}

&&AWGN~\cite{ADM}&
23.21&
22.27&
21.50&
20.97&
28.41&
26.99&
26.23&
25.40&
25.55&
24.29&
23.64&
23.00\\\cline{3-15}

&&Proposed&
\bf 24.08&
\bf 23.09&
\bf 22.29&
\bf 21.69&
\bf 28.70&
\bf 27.34&
\bf 26.51&
\bf 25.70&
\bf 26.03&
\bf 24.80&
\bf 24.12&
\bf 23.42\\\hline\hline
\multirow{3}{*}{$1024$}&
\multirow{3}{*}{$10^{-1}$}&
Input&
17.23&
16.69&
16.24&
15.85&
18.54&
18.19&
17.87&
17.57&
18.27&
17.89&
17.61&
17.38\\\cline{3-15}

&&AWGN~\cite{ADM}&
21.05&
20.18&
19.41&
18.73&
25.57&
24.53&
23.67&
22.95&
23.16&
22.42&
21.86&
21.41\\\cline{3-15}

&&Proposed&
\bf 21.55&
\bf 20.60&
\bf 19.82&
\bf 19.11&
\bf 26.11&
\bf 24.96&
\bf 24.06&
\bf 23.22&
\bf 23.57&
\bf 22.73&
\bf 22.10&
\bf 21.57\\\hline\hline

\multirow{3}{*}{$1024$}&
\multirow{2}{*}{~$10^{-4}$}&
Input&
20.55&
19.46&
18.64&
17.97&
25.02&
23.61&
22.58&
21.76&
23.11&
22.05&
21.34&
20.81\\\cline{3-15}

&&AWGN~\cite{ADM}&
23.21&
22.26&
21.50&
20.96&
28.41&
26.99&
26.23&
25.39&
25.55&
24.29&
23.64&
23.00\\\cline{3-15}

&+Q&Proposed&
\bf 24.08&
\bf 23.08&
\bf 22.28&
\bf 21.68&
\bf 28.70&
\bf 27.34&
\bf 26.50&
\bf 25.70&
\bf 26.02&
\bf 24.79&
\bf 24.11&
\bf 23.41\\\hline\hline
\multirow{3}{*}{$1024$}&
\multirow{2}{*}{~$10^{-1}$}&
Input&
17.38&
16.81&
16.35&
15.94&
18.58&
18.21&
17.89&
17.58&
18.33&
17.93&
17.64&
17.41\\\cline{3-15}

&&AWGN~\cite{ADM}&
20.97&
20.13&
19.38&
18.70&
25.56&
24.53&
23.66&
22.94&
23.14&
22.40&
21.85&
21.40\\\cline{3-15}

&+Q&Proposed&
\bf 21.50&
\bf 20.56&
\bf 19.80&
\bf 19.09&
\bf 26.10&
\bf 24.96&
\bf 24.06&
\bf 23.22&
\bf 23.56&
\bf 22.72&
\bf 22.09&
\bf 21.56\\\hline\hline

\multirow{3}{*}{$256$}&
\multirow{3}{*}{$10^{-4}$}&
Input&
19.93&
18.97&
18.23&
17.62&
23.29&
22.30&
21.51&
20.86&
21.96&
21.13&
20.54&
20.10\\\cline{3-15}

&&AWGN~\cite{ADM}&
22.20&
21.26&
20.50&
19.87&
27.26&
25.90&
25.05&
24.18&
24.48&
23.40&
22.73&
22.18\\\cline{3-15}

&&Proposed&
\bf 22.88&
\bf 21.93&
\bf 21.12&
\bf 20.51&
\bf 27.64&
\bf 26.28&
\bf 25.50&
\bf 24.60&
\bf 24.95&
\bf 23.76&
\bf 23.05&
\bf 22.43\\\hline\hline
\multirow{3}{*}{$256$}&
\multirow{3}{*}{$10^{-1}$}&
Input&
16.93&
16.42&
15.99&
15.62&
18.09&
17.76&
17.47&
17.20&
17.86&
17.52&
17.26&
17.04\\\cline{3-15}

&&AWGN~\cite{ADM}&
20.96&
20.09&
19.32&
18.65&
25.42&
24.39&
23.55&
22.84&
23.03&
22.33&
21.79&
21.34\\\cline{3-15}

&&Proposed&
\bf 21.47&
\bf 20.52&
\bf 19.75&
\bf 19.04&
\bf 25.99&
\bf 24.85&
\bf 23.96&
\bf 23.12&
\bf 23.46&
\bf 22.64&
\bf 22.01&
\bf 21.51\\\hline\hline

\multirow{3}{*}{$256$}&
\multirow{2}{*}{~$10^{-4}$}&
Input&
19.93&
18.97&
18.23&
17.62&
23.29&
22.29&
21.51&
20.85&
21.96&
21.12&
20.54&
20.09\\\cline{3-15}

&&AWGN~\cite{ADM}&
22.19&
21.26&
20.50&
19.86&
27.26&
25.89&
25.05&
24.18&
24.47&
23.40&
22.73&
22.18\\\cline{3-15}

&+Q&Proposed&
\bf 22.88&
\bf 21.93&
\bf 21.12&
\bf 20.51&
\bf 27.64&
\bf 26.28&
\bf 25.50&
\bf 24.60&
\bf 24.95&
\bf 23.76&
\bf 23.05&
\bf 22.43\\\hline\hline
\multirow{3}{*}{$256$}&
\multirow{2}{*}{~$10^{-1}$}&
Input&
17.08&
16.54&
16.10&
15.71&
18.14&
17.80&
17.50&
17.23&
17.92&
17.56&
17.30&
17.07\\\cline{3-15}

&&AWGN~\cite{ADM}&
20.88&
20.04&
19.29&
18.62&
25.40&
24.38&
23.55&
22.84&
23.01&
22.31&
21.77&
21.33\\\cline{3-15}

&+Q&Proposed&
\bf 21.41&
\bf 20.47&
\bf 19.72&
\bf 19.01&
\bf 25.98&
\bf 24.84&
\bf 23.95&
\bf 23.12&
\bf 23.45&
\bf 22.63&
\bf 22.00&
\bf 21.50\\\hline\hline

\multirow{3}{*}{$64$}&
\multirow{3}{*}{$10^{-4}$}&
Input&
18.06&
17.43&
16.90&
16.45&
19.65&
19.19&
18.79&
18.42&
19.09&
18.64&
18.31&
18.03\\\cline{3-15}

&&AWGN~\cite{ADM}&
21.35&
20.45&
19.68&
18.99&
25.95&
24.83&
23.96&
23.18&
23.41&
22.61&
22.02&
21.56\\\cline{3-15}

&&Proposed&
\bf 21.98&
\bf 20.99&
\bf 20.20&
\bf 19.58&
\bf 26.51&
\bf 25.31&
\bf 24.44&
\bf 23.56&
\bf 23.87&
\bf 22.95&
\bf 22.31&
\bf 21.77\\\hline\hline
\multirow{3}{*}{$64$}&
\multirow{3}{*}{$10^{-1}$}&
Input&
15.88&
15.49&
15.15&
14.84&
16.64&
16.40&
16.19&
15.98&
16.51&
16.26&
16.06&
15.90\\\cline{3-15}

&&AWGN~\cite{ADM}&
20.63&
19.81&
19.04&
18.43&
24.88&
23.95&
23.16&
22.54&
22.64&
22.01&
21.53&
21.11\\\cline{3-15}

&&Proposed&
\bf 21.19&
\bf 20.29&
\bf 19.50&
\bf 18.79&
\bf 25.55&
\bf 24.48&
\bf 23.58&
\bf 22.86&
\bf 23.11&
\bf 22.36&
\bf 21.79&
\bf 21.33\\\hline\hline

\multirow{3}{*}{$16$}&
\multirow{3}{*}{$10^{-4}$}&
Input&
14.26&
13.99&
13.77&
13.55&
14.50&
14.35&
14.21&
14.08&
14.42&
14.25&
14.13&
14.02\\\cline{3-15}

&&AWGN~\cite{ADM}&
20.10&
19.31&
18.62&
18.15&
24.02&
23.20&
22.55&
21.99&
21.99&
21.50&
21.08&
20.67\\\cline{3-15}

&&Proposed&
\bf 20.87&
\bf 19.97&
\bf 19.22&
\bf 18.54&
\bf 24.90&
\bf 23.90&
\bf 23.05&
\bf 22.40&
\bf 22.58&
\bf 21.93&
\bf 21.45&
\bf 21.01\\\hline\hline
\multirow{3}{*}{$16$}&
\multirow{3}{*}{$10^{-1}$}&
Input&
13.22&
13.01&
12.83&
12.66&
13.33&
13.23&
13.12&
13.02&
13.36&
13.23&
13.13&
13.04\\\cline{3-15}

&&AWGN~\cite{ADM}&
19.73&
18.97&
18.41&
18.03&
23.48&
22.76&
22.18&
21.65&
21.64&
21.20&
20.81&
20.40\\\cline{3-15}

&&Proposed&
\bf 20.45&
\bf 19.60&
\bf 18.84&
\bf 18.26&
\bf 24.35&
\bf 23.42&
\bf 22.67&
\bf 22.06&
\bf 22.20&
\bf 21.63&
\bf 21.20&
\bf 20.76\\\hline\hline

\multirow{3}{*}{$4$}&
\multirow{3}{*}{$10^{-4}$}&
Input&
9.02&
8.97&
8.92&
8.87&
8.72&
8.69&
8.65&
8.60&
8.81&
8.77&
8.74&
8.70\\\cline{3-15}

&&AWGN~\cite{ADM}&
18.31&
17.99&
17.71&
17.41&
20.86&
20.45&
20.07&
19.67&
19.85&
19.57&
19.26&
18.99\\\cline{3-15}

&&Proposed&
\bf 18.86&
\bf 18.38&
\bf 18.08&
\bf 17.74&
\bf 21.70&
\bf 21.15&
\bf 20.72&
\bf 20.25&
\bf 20.53&
\bf 20.18&
\bf 19.79&
\bf 19.43\\\hline\hline
\multirow{3}{*}{$4$}&
\multirow{3}{*}{$10^{-1}$}&
Input&
8.69&
8.64&
8.60&
8.55&
8.39&
8.35&
8.32&
8.28&
8.49&
8.45&
8.42&
8.39\\\cline{3-15}

&&AWGN~\cite{ADM}&
18.23&
17.93&
17.65&
17.36&
20.64&
20.24&
19.87&
19.49&
19.70&
19.43&
19.14&
18.89\\\cline{3-15}

&&Proposed&
\bf 18.62&
\bf 18.22&
\bf 17.92&
\bf 17.58&
\bf 21.34&
\bf 20.83&
\bf 20.41&
\bf 19.96&
\bf 20.26&
\bf 19.94&
\bf 19.56&
\bf 19.24\\\hline

  \end{tabular}
}
\end{table*}

\begin{figure}[!t]
  \centering
\onecm{
  \begin{tabular}{ccc}
  \includegraphics[width=0.185\textwidth]{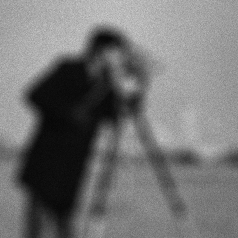}&
  \includegraphics[width=0.185\textwidth]{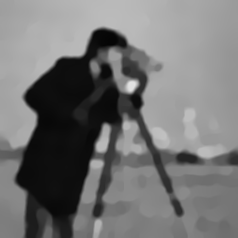}&
  \includegraphics[width=0.185\textwidth]{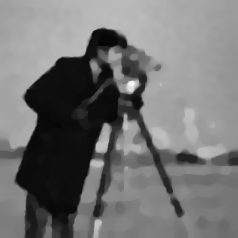}\vspace{-0.8em}\\
  \footnotesize Input (PSNR=18.64dB)&
  \footnotesize AWGN Method~\cite{ADM} (21.48 dB)& 
  \footnotesize Prop. Method (22.29 dB)
  \end{tabular}
}
\twocm{
    \includegraphics[width=0.185\textwidth]{figs/cman_inp.png}\\
    {\footnotesize Input (PSNR=18.64dB)}\\~

    \begin{tabular}{cc}
      \includegraphics[width=0.185\textwidth]{figs/cman_awgn.png}&
      \includegraphics[width=0.185\textwidth]{figs/cman_prop.png}\\
      \footnotesize AWGN Method~\cite{ADM} (21.48 dB)& 
      \footnotesize Prop. Method (22.29 dB)
  \end{tabular}
}
  \caption{Deconvolution results on the Camerman image, blurred with a circular kernel of radius $9$ pixels with noise parameters $\alpha=1024, \sigma=10^{-4}, q = 1/256,$ and $g=2.2$. Note that the proposed method correctly accounts for the noise variance being lower at darker pixels, and recovers sharper estimates in those regions in comparison to the baseline AWGN method of \cite{ADM}.}
  \label{fig:dcnvex}
\end{figure}

We report performance in terms of PSNR (averaged over ten instantiations of the noise) for the AWGN-based ADM-TV method~\cite{ADM}, and for the proposed method using the ADM-TV method as the baseline. We use the parallel optimization approach as described in the appendix, and use the same baseline parameters ($\beta_\nabla=200,\kappa=8,\epsilon=10^{-5}$) in both cases. We set $\beta_0=16$ for the proposed method, and use the mean noise-variance $\sigma_{\mbox{avg}}^2$ for the AWGN results. We note that the proposed approach offers a distinct advantage over the baseline method with better estimates in all cases. Figure~\ref{fig:dcnvex} shows an example of the observed and restored images, and we see that the proposed method is able to account for the noise variance at darker pixels being lower, yielding sharper estimates in those regions. Our approach typically requires only two to three times as many iterations as the baseline method to converge.

\begin{figure*}[!t]
  \onecm{\newcommand{\wdt}{0.16\textwidth}}
  \twocm{\newcommand{\wdt}{0.16\textwidth}}
  \centering
  \begin{tabular}{cccccc}
    \includegraphics[width=\wdt]{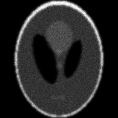}\hspace{-1em}~&
    \includegraphics[width=\wdt]{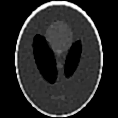}\hspace{-1em}~&
    \includegraphics[width=\wdt]{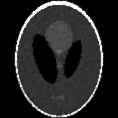}\hspace{-1em}~&
    \includegraphics[width=\wdt]{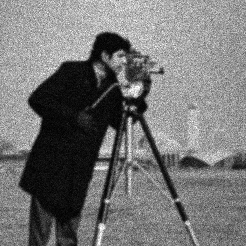}\hspace{-1em}~&
    \includegraphics[width=\wdt]{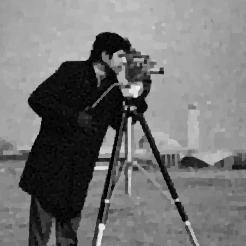}\hspace{-1em}~&
    \includegraphics[width=\wdt]{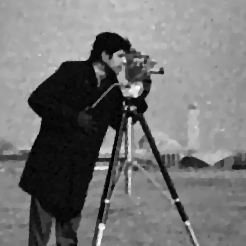}\hspace{-1em}~\onecm{\vspace{-0.5em}}\\
    \footnotesize Input&
    \footnotesize SPIRAL-TV&
    \footnotesize Prop.+TV&
    \footnotesize Input&
    \footnotesize SPIRAL-TV&
    \footnotesize Prop.+TV\onecm{\vspace{-1em}}\\
    \footnotesize PSNR=20.29 dB&
    \footnotesize 23.76 dB (4.2s) &
    \footnotesize 24.70 dB (0.7s)&
    \footnotesize 21.33 dB&
    \footnotesize 25.52 dB (94.0s)&
    \footnotesize 25.60 dB (1.7s)
  \end{tabular}

  \caption{Deconvolution performance with Poisson Noise in comparison to the SPIRAL-TV method~\cite{spiraltv}. The running times are indicated in brackets.}
  \label{fig:spiralcmp}
\end{figure*}

Next, we show results in Fig.~\ref{fig:spiralcmp} for the two demonstration examples provided by authors of \cite{spiraltv}, each blurred with a $5\times 5$ kernel and corrupted only by Poisson noise. We show the running times and PSNR values of the restored images, from both the proposed method (using the same parameters as above), and the SPIRAL technique~\cite{spiraltv} with a TV prior (which yields the highest PSNR values amongst the different priors). In both cases, our approach yields estimates with higher accuracy, and offers a significant advantage in computational cost---  for the \emph{cameraman} example, our method converges in just 25 iterations, while the SPIRAL technique requires a 100 iterations, with each iteration in-turn calling an iterative baseline TV-solver.

\begin{figure}[!t]
  \centering
\onecm{
  \begin{tabular}{cccc}
  \includegraphics[width=0.19\textwidth]{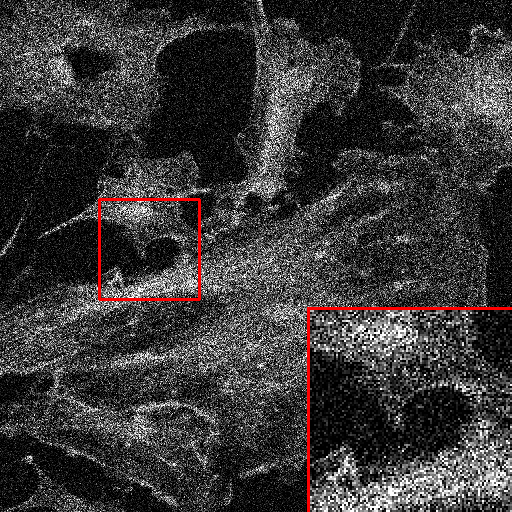}&
  \includegraphics[width=0.19\textwidth]{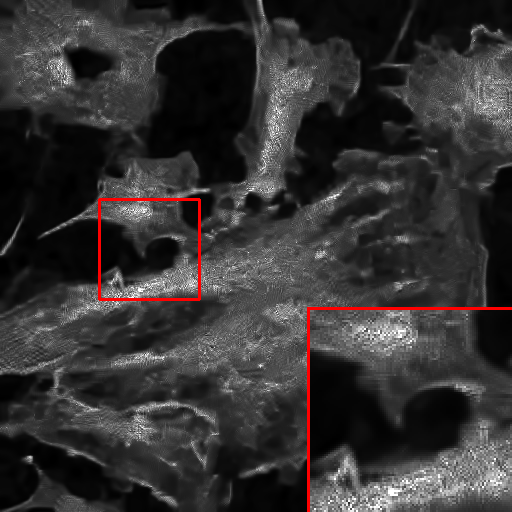}&
  \includegraphics[width=0.19\textwidth]{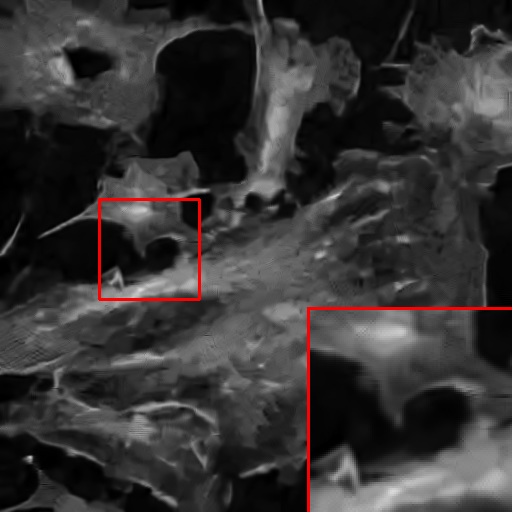}&
  \includegraphics[width=0.19\textwidth]{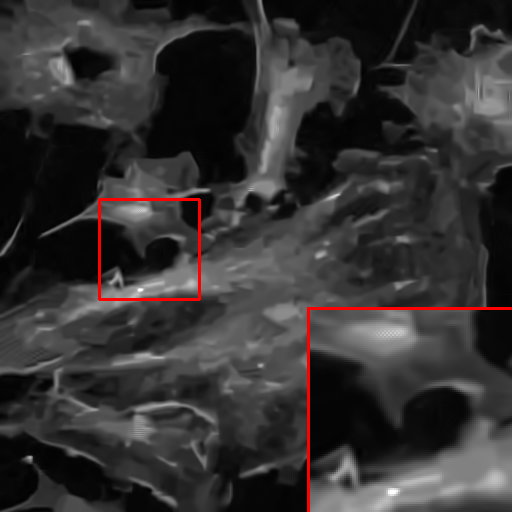}\vspace{-0.8em}\\
  \footnotesize Input (PSNR=13.40dB)&
  \footnotesize BM3D~\cite{bm3d} (25.54 dB)& 
  \footnotesize GAT+BM3D~\cite{foi} (27.57 dB)& 
  \footnotesize Prop.+BM3D (27.79 dB)
  \end{tabular}
}
\twocm{
    \begin{tabular}{cc}
      \hspace{-0.5em}\includegraphics[width=0.245\textwidth]{figs/fc_nz_inp_cu.png}\hspace{-0.8em}~&
      \hspace{-0.5em}\includegraphics[width=0.245\textwidth]{figs/fc_nz_awgn_cu.png}\hspace{-0.8em}~\\
      \footnotesize Input (PSNR=13.40dB) & 
      \footnotesize BM3D~\cite{bm3d} (25.54 dB)\\\vspace{-0.25em}~\\
      \hspace{-0.5em}\includegraphics[width=0.245\textwidth]{figs/fc_nz_foi_cu.png}\hspace{-0.8em}~&
      \hspace{-0.5em}\includegraphics[width=0.245\textwidth]{figs/fc_nz_prop_cu.png}\hspace{-0.8em}~\\
      \footnotesize GAT+BM3D~\cite{foi} (27.57 dB)& 
      \footnotesize Prop.+BM3D (27.79 dB)
  \end{tabular}
}
  \caption{Denoising results (with inset close-ups) on the Fluorescent cells image, with Poisson-Gaussian noise corresponding to a peak input intensity of $5$ and $\sigma=0.5$.}
  \label{fig:dnzex}
\end{figure}

Finally, we report denoising performance (i.e., $k=\delta$) in Table.~\ref{tab:dnz}, for the standard test cases used in \cite{foi} with Poisson-Gaussian noise. The state-of-the-art AWGN techniques for denoising tend to be more sophisticated  than those for deconvolution. We use the BM3D~\cite{bm3d} algorithm, which uses a complex adaptive image prior, as the baseline estimator for our approach in this case (with $\beta_0=2,\epsilon=10^{-3}$). In addition to PSNR values for the proposed method, we show results for the baseline AWGN method (with $\sigma_{\mbox{avg}}^2$ as the noise variance), and for the Poisson-Gaussian denoising algorithms described in \cite{foi} and \cite{florian}. Note that the method in \cite{foi} also uses BM3D as a baseline. We use the same notation as in \cite{foi} to describe the noise parameters in Table~\ref{tab:dnz}, where noise is synthetically added by first scaling the input image to a certain peak value, instantiating Poisson random variables with these scaled intensities, and then adding Gaussian noise with variance $\sigma^2$. The reported PSNR values are again averaged over ten instantiations of the noise. Figure \ref{fig:dnzex} shows an example of input and restored images for this case.
\begin{table*}[!t]
\twocm{\renewcommand{\arraystretch}{1.3}}
\onecm{\renewcommand{\arraystretch}{0.9}}
  \centering
  \caption{Denoising performance (PSNR in dB) with Poisson-Gaussian Noise}
  \label{tab:dnz}
{\onecm{\tiny}
  \begin{tabular}{|l||c|c|c||c||c|c||c|c||c|}
    \hline
    Image & Peak & $\sigma$ & Noisy & ~~BM3D~\cite{bm3d}~~ & GAT+BM3D~\cite{foi} & PURE-LET~\cite{florian} & Prop.+BM3D & \# Iterations & Prop.+TV\\\hline\hline

    \multirow{8}{*}{Cameraman} 
    & 1 & 0.1 & 3.20 & 18.50 & 20.23 & 20.35 &\bf  20.71& 7.2&17.42\\\cline{2-10}
    & 2 & 0.2 & 6.12 & 20.95 & 21.93 & 21.60 &\bf  22.12& 5.1&18.53\\\cline{2-10}
    & 5 & 0.5 & 9.83 & 23.55 & 24.09 & 23.33 &\bf 24.10& 4.0&21.08\\\cline{2-10}
    & 10 & 1 & 12.45 & 25.10 & 25.52 & 24.68 &\bf 25.57& 4.0&22.88\\\cline{2-10}
    & 20 & 2 & 14.76 & 26.50 & 26.77 & 25.92 &\bf  26.81& 4.0&24.55\\\cline{2-10}
    & 30 & 3 & 15.91 & 27.10 &\bf 27.30 & 26.51 &\bf  27.30& 4.0& 25.26\\\cline{2-10}
    & 60 & 6 & 17.49 & 27.97 & 28.07 & 27.35 &\bf  28.10& 3.0&26.12\\\cline{2-10}
    & 120 & 12 & 18.57 & 28.52 & 28.57 & 27.89 &\bf 28.61& 3.0&26.56\\\hline\hline

    \multirow{8}{*}{\parbox[c]{0.07\textwidth}{
        Fluorescent\\Cells}
    } 
    & 1 & 0.1 & 7.22 & 19.64 & 24.54 & \bf 25.13 & 24.91& 8.8&20.79\\\cline{2-10}
    & 2 & 0.2 & 9.99 & 22.24 & 25.87 & \bf 26.25 & 26.16& 7.3&22.68\\\cline{2-10}
    & 5 & 0.5 &13.37 & 25.43 & 27.45 & 27.60 &\bf 27.72& 5.5&25.21\\\cline{2-10}
    & 10 & 1 & 15.53 & 27.53 & 28.63 & 28.59 &\bf 28.77& 5.0&27.11\\\cline{2-10}
    & 20 & 2 & 17.21 & 29.18 & 29.65 & 29.47 &\bf 29.69& 5.0&28.37\\\cline{2-10}
    & 30 & 3 & 17.97 & 29.91 & 30.16 & 29.84 &\bf 30.18& 4.0&28.87\\\cline{2-10}
    & 60 & 6 & 18.86 & 30.72 & \bf 30.77 & 30.42 & 30.75& 4.0&29.38\\\cline{2-10}
    & 120 & 12 & 19.39 & \bf 31.15 & 31.14 & 30.70 & 31.09& 4.0&29.60\\\hline\hline

    \multirow{8}{*}{Lena} 
    & 1 & 0.1 & 2.87 & 19.87 & 22.59 & \bf 22.83 & 22.58& 7.0&17.01\\\cline{2-10}
    & 2 & 0.2 & 5.82 & 22.58 &\bf 24.34 & 24.16 & 24.14& 5.1&19.68\\\cline{2-10}
    & 5 & 0.5 & 9.54 & 25.42 &26.17 & 25.74 &\bf 26.19& 4.0&23.18\\\cline{2-10}
    & 10 & 1 & 12.19 & 27.20 &27.72 & 27.27 &\bf 27.80& 4.0&25.41\\\cline{2-10}
    & 20 & 2 & 14.53 & 28.66 &29.01 & 28.46 &\bf 29.12& 4.0&27.05\\\cline{2-10}
    & 30 & 3 & 15.72 & 29.42 &29.69 & 29.12 &\bf 29.75& 3.5&27.69\\\cline{2-10}
    & 60 & 6 & 17.35 & 30.37 & 30.51 & 29.91 &\bf 30.57& 3.0&28.12\\\cline{2-10}
    & 120 & 12 & 18.48 & 30.98 & 31.05 & 30.51 &\bf 31.11& 3.0&28.08\\\hline

  \end{tabular}
}
\end{table*}

We find that the proposed approach is competitive with these existing methods, with the highest PSNR in a majority of the test cases. However, it is important to remember that our approach is iterative, while the algorithms in \cite{foi,florian} are single-shot. Table.~\ref{tab:dnz} reports the mean number of iterations required in each case, and we see that convergence is usually quick, requiring roughly three to seven calls to the baseline method. We also show results for denoising using the simple TV prior (which was used for the deconvolution results in Table.~\ref{tab:dcnv}), and note that using the BM3D method as the baseline instead leads to a significant improvement. This highlights the importance of the flexibility that our approach offers, in allowing the use of any baseline AWGN restoration method rather than being tied to a fixed prior.

\section{Conclusion}
\label{sec:conc}

In this paper, we introduce a framework for image restoration in the presence of signal-dependent noise. We describe an observation model that accounts for camera sensor measurements being corrupted by both Gaussian and signal-dependent Poisson noise sources, and for errors from the subsequent digitization of these measurements. Then, we use \emph{variable-splitting} to derive a fast iterative scheme that is able to adapt existing AWGN-based restoration methods for inference with this noise model. A MATLAB implementation of the algorithm, along with scripts to generate the results presented in this paper, is available for download at \url{http://vision.seas.harvard.edu/PGQrestore/}.

The flexibility in being able to incorporate any AWGN-based method as a baseline means that we can draw on a considerable amount of existing research on image statistics. Our approach can therefore be easily used for restoration of any class of images (such as medical images, or astronomical data) which is corrupted by noise with similar characteristics, by using an appropriate class-specific AWGN-restoration technique as the baseline. Moreover, the optimization scheme described here can be adapted to other noise models, as long as the likelihood functions for those models are convex, or can be so approximated.

\appendix

The ADM-TV method~\cite{ADM} is a popular choice for deconvolution under AWGN, and is itself based on variable-splitting. When using this method as the baseline algorithm, it is possible to adopt an approach where its internal optimization is effectively done in parallel to that for the noise model in our framework. We describe this approach in detail in this appendix. The ADM-TV method uses an image prior that penalizes the TV-norm of image gradients:
\begin{equation}
  \label{eq:TVL1}
  \Phi(x) = \kappa \sum_n \sqrt{\sum_i |(x*\nabla_i)(n)|^2},
\end{equation}
where $\nabla_i$ are gradient filters, and $\kappa$ is a model parameter. The MAP estimate in this case is computed by introducing additional auxiliary variables $d_i(n)$ corresponding to the image gradients, and the estimation problem is cast as:
\begin{eqnarray}
  \label{eq:adm}
  x = \arg \min_x \kappa \sum_{n} \sqrt{\sum_i|d_i(n)|^2} + L(\tau(n);y(n)),\notag\\
  \mbox{subject to:~} d_i(n) = (x*\nabla_i)(n);~~\tau(n) = (x*k)(n).
\end{eqnarray}
We define a \emph{joint} augmented Lagrangian-based cost for this case as in \cite{uclatr}:
\begin{eqnarray}
  \label{eq:alm2}
C(x,d_i,\tau,\lambda) \onecm{&}=\onecm{&}\kappa\sum_{n} \sqrt{\sum_i |d_i(n)|^2} + \left[  \frac{\beta_\nabla}{2} \sum_{n,i} (d_i(n)-x_i(n))^2 
\twocm{\right.\hspace{-2em}\notag \\\left.} 
- \sum_{n,i} \lambda_i(n) (d_i(n) - x_i(n)) \right] 
\onecm{\notag\\&&}
+ \left[ \frac{\beta}{2} \sum_n (\tau(n)-x_k(n))^2 
\twocm{\right.\hspace{-2em}\notag\\\left.}
-\sum_n \lambda(n) (\tau(n) - x_k(n)) \right] + L(\tau(n);y(n)),~
\end{eqnarray}
where $x_i(n) = (x*\nabla_i)(n)$, and the second and third term above encode the gradient equality constraint in \eqref{eq:adm}.

The iterative optimization for this case proceeds as follows:
\begin{eqnarray}
  \label{eq:cupdate0}
  d_i^{t+1} &\leftarrow& \arg \min_{d_i} C(x^t,d_i,\tau^t,\lambda^t),\\
  \label{eq:cupdate1}
  x^{t+1} &\leftarrow& \arg \min_x C(x,d_i^{t+1},\tau^t,\lambda^t),\\
  \label{eq:cupdate2}
  \tau^{t+1} &\leftarrow& \arg \min_\tau C(x^{t+1},d_i^{t+1},\tau,\lambda^t),\\
  \label{eq:cupdate3}
  \lambda^{t+1}(n) &\leftarrow& \lambda^t(n) - \gamma(n) \beta (\tau^{t+1}(n)-x_k^{t+1}(n)),\\
  \label{eq:cupdate4}
  \lambda_i^{t+1}(n) &\leftarrow& \lambda^t_i(n) - \gamma_{\mbox{\tiny max}} \beta_\nabla (d_i^{t+1}(n)-x_i^{t+1}(n)).
\end{eqnarray}
Note that this is essentially equivalent to the optimization framework in \eqref{eq:update1}-\eqref{eq:update3}, with \eqref{eq:cupdate0},~\eqref{eq:cupdate1} corresponding to the $x$ update step in \eqref{eq:update1}, and \eqref{eq:cupdate4} being an extra update step for the additional Lagrange multipliers (using a fixed step size equal to $\gamma_{\mbox{\tiny max}}$, as in \cite{ADM}). The solution to the $\tau$ update step is identical to the one described in \eqref{eq:tausol} in Sec.~\ref{sec:algo}, since the minimization in \eqref{eq:cupdate2} for this update depends only on the last three terms in the cost in \eqref{eq:alm2}.

The update steps to $d_i$ and $x$ in \eqref{eq:cupdate0}, \eqref{eq:cupdate1} are largely independent of the noise model, and are similar to those described in \cite{ADM}. Specifically, each $d_i^{t+1}(n)$ can be computed independently on a per-pixel basis:
\begin{eqnarray}
  \label{eq:eqdiupd}
  d_i^{t+1}(n)\hspace{-0.5em}&=\hspace{-0.5em}& \max\left(0,\sqrt{\tilde{d}(n)} -\beta_\nabla^{-1}\kappa\right) \frac{\left(x_i^t(n)+\beta_\nabla^{-1}\lambda_i^t(n) \right)}{{\sqrt{\tilde{d}(n)}}},\twocm{\notag\\}
~~\tilde{d}(n)\twocm{\hspace{-0.5em}&}=\twocm{\hspace{-0.5em}&}  \sum_i |x_i(n)+\beta_\nabla^{-1}\lambda_i^t(n)|^2.
\end{eqnarray}
The updated value of $x$ can then be computed efficiently in the Fourier domain as:
\begin{equation}
  \label{eq:xtvupd}
  x^{t+1} = \mathcal{F}^{-1}\left[ \frac{\beta_\nabla\sum_i \mathcal{F}[\nabla_i]^{^*}\mathcal{F}[\tilde{d}_i(n)] + \beta \mathcal{F}[k]^{^*}\mathcal{F}[\tilde{\tau}(n)] }{\beta_\nabla \sum_i |\mathcal{F}[\nabla_i]|^2 + \beta |\mathcal{F}[k] |^2}\right],~~~
\end{equation}
where,
\begin{equation}
  \label{eq:xxd}
  \tilde{d}_i(n)  = d_i^{t+1}(n) - \beta_\nabla^{-1}\lambda_i^t(n),~~
  \tilde{\tau}(n) = \tau^{t}(n) - \beta^{-1}\lambda^t(n),
\end{equation}
\twocm{\vfill\pagebreak\noindent}%
and $\mathcal{F}$ and $\mathcal{F}^{-1}$ refer to the forward and inverse discrete Fourier transforms. To account for the periodicity assumption with the Fourier transform in \eqref{eq:xtvupd}, we extend $\tilde{d}_i$ and $\tilde{\tau}$ in each direction by six times the size of the kernel $k$ before computing the Fourier transform, with zeros for $\tilde{d}_i$, and values that linearly blend the intensities at opposite boundaries for $\tilde{\tau}$. After computing $x^{t+1},x_k^{t+1}$ and $x_i^{t+1}$, we crop them back to their original sizes.

We begin the iterations with $\tau(n)=m_q(y(n)),~\lambda(n)=0,~x(n)=0$ and $\lambda_i(n) = 0$, and choose the gradient filters $\nabla_i$ to correspond to horizontal and vertical finite-differences (i.e., $[-1,1]$). We vary the step sizes $\gamma(n)$ as described in Sec.~\ref{sec:algo}. Moreover, we do not apply 
the updates in \eqref{eq:cupdate2} and \eqref{eq:cupdate4} to $\tau(n)$ and $\lambda(n)$ for the first few iterations (six in our implementation), during which time the algorithm proceeds identically to the AWGN ADM-TV~\cite{ADM} method with input $m_q(y(n))$ and noise variance $\beta^{-1}$.

\bibliographystyle{IEEEtran}
\bibliography{realnoise}

\begin{thebibliography}{10}
\providecommand{\url}[1]{#1}
\csname url@samestyle\endcsname
\providecommand{\newblock}{\relax}
\providecommand{\bibinfo}[2]{#2}
\providecommand{\BIBentrySTDinterwordspacing}{\spaceskip=0pt\relax}
\providecommand{\BIBentryALTinterwordstretchfactor}{4}
\providecommand{\BIBentryALTinterwordspacing}{\spaceskip=\fontdimen2\font plus
\BIBentryALTinterwordstretchfactor\fontdimen3\font minus
  \fontdimen4\font\relax}
\providecommand{\BIBforeignlanguage}[2]{{%
\expandafter\ifx\csname l@#1\endcsname\relax
\typeout{** WARNING: IEEEtran.bst: No hyphenation pattern has been}%
\typeout{** loaded for the language `#1'. Using the pattern for}%
\typeout{** the default language instead.}%
\else
\language=\csname l@#1\endcsname
\fi
#2}}
\providecommand{\BIBdecl}{\relax}
\BIBdecl

\bibitem{portilla}
J.~Portilla, V.~Strela, M.~Wainwright, and E.~Simoncelli, ``Image denoising
  using scale mixtures of {G}aussians in the wavelet domain,'' \emph{IEEE
  Trans. Imag. Proc.}, 2003.

\bibitem{bm3d}
K.~Dabov, A.~Foi, V.~Katkovnik, and K.~Egiazarian, ``Image denoising by sparse
  {3D} transform-domain collaborative filtering,'' \emph{IEEE Trans. Imag.
  Proc.}, 2008.

\bibitem{levin07}
A.~Levin, R.~Fergus, F.~Durand, and W.~T. Freeman, ``Image and depth from a
  conventional camera with a coded aperture,'' in \emph{ACM TOG (SIGGRAPH)},
  2007.

\bibitem{ADM}
M.~Tao and J.~Yang, ``Alternating direction algorithms for total variation
  deconvolution in image reconstruction,'' Department of Mathmatics, Nanjing
  University, Tech. Rep. TR0918, 2009.

\bibitem{dilip}
D.~Krishnan and R.~Fergus, ``Fast image deconvolution using {H}yper-{L}aplacian
  priors,'' in \emph{NIPS}, 2009.

\bibitem{kolaczyk}
E.~Kolaczyk, ``Bayesian multiscale models for {P}oisson processes,'' \emph{J.
  Am. Statist. Ass.}, 1999.

\bibitem{hirakawa}
K.~Hirakawa and P.~Wolfe, ``Efficient multivariate skellam shrinkage for
  denoising photon-limited image data: An empirical bayes approach,'' in
  \emph{Proc.~ICIP}, 2009.

\bibitem{florian}
F.~Luisier, T.~Blu, and M.~Unser, ``Image denoising in mixed poisson-gaussian
  noise,'' \emph{IEEE Trans. Imag. Proc.}, 2011.

\bibitem{foi0}
M.~M\"{a}kitalo and A.~Foi, ``Optimal inversion of the {A}nscombe
  transformation in low-count {P}oisson image denoising,'' \emph{IEEE Trans.
  Imag. Proc.}, 2011.

\bibitem{foi}
------, ``Optimal inversion of the generalized {A}nscombe transformation for
  {P}oisson-{G}aussian noise,'' \emph{IEEE Trans. Imag. Proc.}, draft accessed
  from http://www.cs.tut.fi/\~{}foi/publications.html on 13 March, 2012.

\bibitem{spiraltv}
Z.~Harmany, R.~Marcia, and R.~Willett, ``This is {SPIRAL-TAP}: {S}parse
  {P}oisson {I}ntensity {R}econstruction {AL}gorithms {---} {T}heory and
  {P}ractice,'' \emph{IEEE Trans. Imag. Proc.}, 2012.

\bibitem{geman}
D.~Geman and C.~Yang, ``Nonlinear image recovery with half-quadratic
  regularization,'' \emph{IEEE Trans. Imag. Proc.}, 1995.

\bibitem{uclatr}
C.~Wu, J.~Zhang, and X.~Tai, ``Augmented lagrangian method for total variation
  restoration with non-quadratic fidelity,'' \emph{Inverse Problems and
  Imaging}, 2011.

\end{thebibliography}

\end{document}